\documentclass[lettersize,journal]{IEEEtran}  % Comment this line out if you need a4paper

\IEEEoverridecommandlockouts                              % This command is only needed if 
\usepackage{verbatim}
\usepackage{amsmath}

\usepackage[utf8]{inputenc}
\usepackage{graphicx}
\usepackage{subfigure}
\usepackage{float}
\usepackage{bm}
\usepackage{amssymb}
\usepackage{tcolorbox}
\usepackage{xcolor}
\usepackage{threeparttable}
\usepackage{multirow}
\usepackage{booktabs}
\usepackage{tabularray}

\usepackage{graphics} % for pdf, bitmapped graphics files
\usepackage{epsfig} % for postscript graphics files
\usepackage[T1]{fontenc}
\usepackage{cite}
\usepackage{tabularx}
\usepackage{algorithm}
\usepackage{algorithmic}
\usepackage{color}
\usepackage{hyperref}
\hypersetup{colorlinks,allcolors=black}

\graphicspath{{Images/}}
\title{\LARGE \bf
A Semantic-Aware Framework for Safe and Intent-Integrative Assistance in Upper-Limb Exoskeletons
}
\author{Yu Chen, Shu Miao, Chunyu Wu, Jingsong Mu, Bo OuYang, and Xiang Li% <-this % stops a space
\thanks{Y. Chen, S. Miao, and X. Li are with the Department of Automation, Tsinghua University, China. This work was supported in part by the Science and Technology Innovation 2030-Key Project under Grant 2021ZD0201404, in part by the National Natural Science Foundation of China under Grant U21A20517 and 62461160307, in part by the Beijing National Research Center for Information Science and Technology, and in part by China Postdoctoral Science Foundation 2025M771713. 
Corresponding author: Xiang Li (xiangli@tsinghua.edu.cn)}%
}

\begin{document}

\maketitle
\pagestyle{plain}  % no page number for the second and the later pages
\thispagestyle{plain} % no page number for the first page
%\thispagestyle{plain}
%\pagestyle{plain}

%%%%%%%%%%%%%%%%%%%%%%%%%%%%%%%%%%%%%%%%%%%%%%%%%%%%%%%%%%%%%%%%%%%%%%%%%%%%%%%%
\begin{abstract}
% Upper-limb exoskeleton robots are commonly developed for rehabilitation of daily life activities. Specifically, this type of robots enhance the manipulability of wearer over the training process.
% Different from the lower-limb exoskeletons, which are employed for weight support and rhythmic assistance, upper-limb exoskeleton robots are facing dexterous manipulation task (i.e., pick and place), requiring expressive instruction setting and comprehensive assistance.
Upper-limb exoskeletons are primarily designed to provide assistive support by accurately interpreting and responding to human intentions.
In home-care scenarios, exoskeletons are expected to adapt their assistive configurations based on the semantic information of the task, adjusting appropriately in accordance with the nature of the object being manipulated. 
However, existing solutions often lack the ability to understand task semantics or collaboratively plan actions with the user, limiting their generalizability.
To address this challenge, this paper introduces a semantic-aware framework that integrates large language models into the task planning framework, enabling the delivery of safe and intent-integrative assistance.
The proposed approach begins with the exoskeleton operating in transparent mode to capture the wearer’s intent during object grasping.
Once semantic information is extracted from the task description, the system automatically configures appropriate assistive parameters. In addition, a diffusion-based anomaly detector is used to continuously monitor the state of human-robot interaction and trigger real-time replanning in response to detected anomalies.
During task execution, online trajectory refinement and impedance control are used to ensure safety and regulate human-robot interaction. Experimental results demonstrate that the proposed method effectively aligns with the wearer’s cognition, adapts to semantically varying tasks, and responds reliably to anomalies.
\begin{comment} 
To add experiment illustration
\end{comment}

\end{abstract}

%%%%%%%%%%%%%%%%%%%%%%%%%%%%%%%%%%%%%%%%%%%%%%%%%%%%%%%%%%%%%%%%%%%%%%%%%%%%%%%%
\section{Introduction}

\IEEEPARstart{E}{xoskeleton} robots have emerged as a promising solution to address the increasing challenges associated with aging populations~\cite{grimmer2019mobility}, offering precise assistance that improves users' mobility. Lower-limb exoskeletons, such as ReWalk~\cite{esquenazi2017powered} and HAL~\cite{iizuka2021regulation}, are widely used to support body weight and facilitate various walking tasks~\cite{chen2024learning}. However, their primary focus remains on locomotion. Older individuals receiving home care often require diverse forms of assistance beyond walking, particularly for tasks requiring upper-limb functionality. Upper-limb exoskeletons~\cite{Shu2023TwoStageTC, Kazerooni2008ExoskeletonsFH, Zimmermann2023ANYexo2A} are designed to support the range of motion needed for activities of daily living and can enhance upper-limb functionality, especially in tasks involving repetitive movements.

% using force-sensing resistors to fultill intentional reaching direction\cite{huang2015control}.
% same, using rule to recognize and integrated into admittance control\cite{huo2011control}.
% brain-machine interface to estimate human intention\cite{qiu2016brain}.
% multi-modal and deep learning \cite{sedighi2023emg}.
% using Gaussian radial basis function network \cite{wu2020development}.

% The assistance provided by exoskeletons is often reliant on pre-defined trajectories, which can limit adaptability. 
Exoskeleton robots are developed to assist users in performing various tasks. However, the close and frequent human-robot interaction involved in these tasks makes safety a top priority throughout the entire process.
To ensure safety, several measures have been implemented, both in hardware and software systems. On the hardware side, series elastic actuators (SEAs)\cite{Veneman2007DesignAE} and cable-driven mechanisms\cite{Veneman2006ASE} have been incorporated into exoskeleton designs to enhance compliance and flexibility.
On the software side, advanced control methods, such as backstepping control\cite{Li2017AdaptiveHI} and singular perturbation theory\cite{li2017multi}, are used to achieve accurate position and force regulation, even in systems with higher-order nonlinear dynamics.
Although these developments have improved control precision, existing frameworks still rely on predefined trajectories throughout the assistance process, which limits the adaptability of the system.
To meet the diverse demands of upper-limb motion, incorporating human intention and enabling mode switching across different movement patterns provide a flexible and responsive approach to assistance~\cite{huo2011control}. For instance, force-sensing resistors have been used to accurately determine the direction of a user's intended movement, allowing the exoskeleton to assist in a manner that corresponds to the wearer’s actions~\cite{huang2015control}. In addition, brain-machine interfaces that use electroencephalography signals have been developed to extract user intention directly~\cite{qiu2016brain}.
To improve the accuracy of intention estimation and broaden the applicability of these methods, researchers have explored multi-modal sensor-based approaches. By integrating advanced techniques such as deep learning~\cite{sedighi2023emg} and Gaussian radial basis function networks~\cite{wu2020development}, these systems can effectively recognize diverse motion patterns, paving the way for more personalized and adaptive exoskeleton assistance.
However, existing frameworks remain limited by fixed controller settings and safety constraints~\cite{chen2024safe}, often resulting in overly conservative motion planning that cannot adapt to task-specific requirements. Furthermore, the absence of human intention integration throughout the assistance process reduces the flexibility and generalizability of the system for various tasks.

% Exoskeleton robots require close and frequent human-robot interactions during assistance, making safety a top priority throughout the process. To address this, series elastic actuators (SEAs)~\cite{Veneman2007DesignAE} and cable-driven mechanisms~\cite{Veneman2006ASE} have been incorporated into exoskeleton designs. These mechanical innovations enhance compliance and flexibility by absorbing unexpected impacts and mitigating inertial effects on the limbs during motion.
% On the software side, advanced control methods have been implemented to ensure precise position and force control in systems with higher-order nonlinear dynamics. Techniques such as backstepping control~\cite{Li2017AdaptiveHI} and singular perturbation theory~\cite{li2017multi} have been used to enhance control accuracy. Additionally, iterative learning control \cite{Li2018IterativeLI} and model-based adaptive controllers \cite{wang2022extracting} have been developed for effective disturbance compensation.
% Despite these advancements, current control frameworks are often limited to fixed controller settings or predefined safety constraints \cite{chen2024safe}. As a result, motion planning relies on overly conservative strategies that fail to adapt to task-specific requirements. This lack of flexibility prevents dynamic adjustments to the controller configuration, restricting the system's ability to optimize performance for varying tasks.

To deliver comprehensive assistance to humans, the development of large language models (LLMs) has introduced advanced reasoning capabilities and a deep understanding of semantic information into robotic systems. These models support logical analysis and the generation of step-by-step responses to described tasks~\cite{brohan2023can}. In addition, they can directly produce motion commands from the action space~\cite{zitkovich2023rt}. However, in the context of exoskeleton robots, effectively integrating human intention with planned actions generated by LLMs remains an open challenge.
{Current approaches mainly use LLMs as natural language processing modules to enable voice-based input for motion control~\cite{parameswari2024assessment,rifai2024upper}. For instance, verbal commands can be used to adjust exoskeleton joint angles~\cite{chen2024llm}. Although these methods are functional, they often overlook the rich semantic information embedded in task descriptions conveyed through language, limiting their ability to provide appropriate assistance across diverse manipulated objects.}
Fully leveraging LLMs to interpret and utilize this semantic information can significantly enhance the scope and adaptability of exoskeleton assistance.

To address these challenges, this paper proposes a semantic-aware framework for upper-limb exoskeletons that can assist the wearer in a safe manner, allowing the wearer's motion intentions to intervene and adjust the current assistance strategy during the assistance process.
The contributions are summarized as follows:
\begin{enumerate}

    \item[1)] \emph{LLM Embodied Exoskeleton}: {This work introduces the integration of LLMs with upper-limb exoskeletons by introducing reasoning and planning capabilities that enable the system to comprehend and respond to language-based commands while considering the semantic information.}
    By decomposing assistive tasks into multi-step processes, LLMs improve task success rates and enhance the exoskeleton's generalizability across diverse operations, thereby improving the user-friendliness and interaction quality of the exoskeleton.

    \item[2)] \emph{Intent-Integrative Strategy}: The proposed framework provides the wearer with a high degree of freedom to express motion intentions. By using the exoskeleton's transparent mode, users can independently determine the object to be manipulated, thereby preventing errors such as misgrasping. Assistance is provided only during tasks where it is explicitly required. A novel anomaly detection network ensures that any misaligned movement intentions interrupt the current assistance strategy and trigger a replanning process, enhancing adaptability and precision.

    \item[3)] \emph{Semantic-Aware Assistance}: 
    In addition to reasoning and planning, LLMs are used to extract and interpret semantic information embedded in task descriptions. This capability enables the exoskeleton to account for safety-related attributes in the task environment during assistance. Combined with the proposed impedance controller and online trajectory refinement method, the exoskeleton dynamically adjusts impedance parameters and trajectory constraints based on semantic insights, ensuring controlled, comprehensive, and contextually effective assistance.
\end{enumerate}

The proposed framework enables the exoskeleton robot to set impedance and speed parameters, apply safety constraints, and plan actions based on the task description. Simultaneously, it allows real-time intervention by the wearer to address anomalies that may occur during task execution. The feasibility of this approach has been validated using a cable-driven upper-limb exoskeleton.

\section{System Structure}
% \subsection{Model of Compliantly-Driven Exoskeleton}
% \tcolor{- to add the structure of exoskeleton robots, including mechanical design, and the SEA + cable-driven actuator -}

\begin{figure}[!t]
    \centering
    \includegraphics[width=7cm]{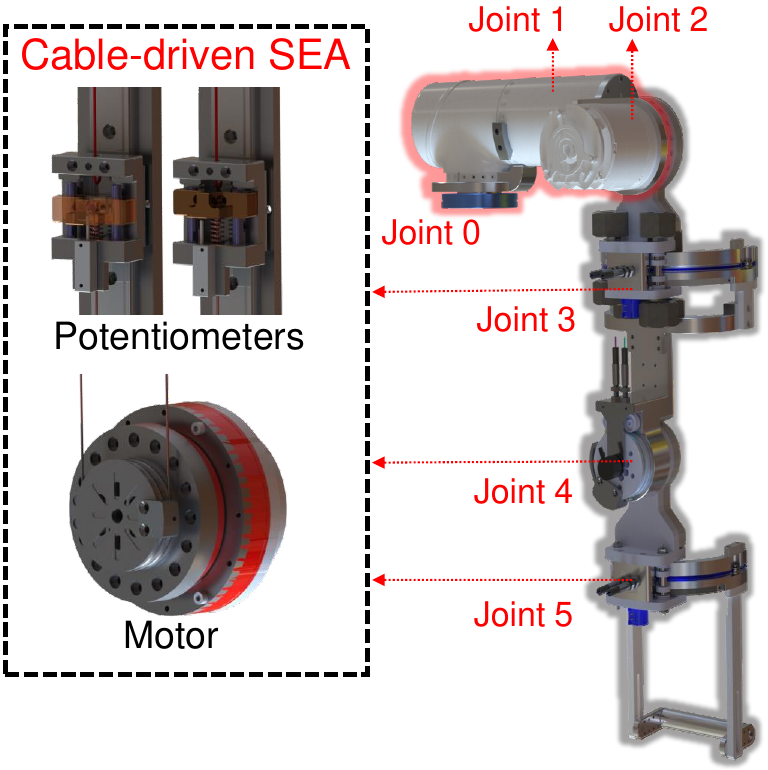}
    \caption{The upper-limb exoskeleton includes one passive joint (Joint 0), two active direct-driven joints (Joint 1 and 2), and three cable-driven joints (Joints 3--5) with each unit composed of a motor and two potentiometers.}
    \label{device}
    % \vspace{-1cm}
\end{figure}

The construction of the upper-limb exoskeleton used in this study is illustrated in Fig.~\ref{device}. To provide compliant and comprehensive assistance, the exoskeleton incorporates SEAs and cable-driven mechanisms, which effectively reduce the impact of inherent inertia during limb movement and enhance interaction quality.
The exoskeleton features five active joints (Joints 1–-5) and one passive joint (Joint 0) to support the range of upper-limb movements required for daily activities in the joint space. The specific functions of these six joints are summarized in Table~\ref{Joints_func}.
% \begin{enumerate}
% \item[-] Joint 0: Shoulder eccentric movement
% \item[-] Joint 1: Shoulder abduction/adduction 
% \item[-] Joint 2: Shoulder flexion/extension 
% \item[-] Joint 3: Upper-arm internal and external rotation
% \item[-] Joint 4: Elbow flexion/extension
% \item[-] Joint 5: Forearm internal and external rotation
% \end{enumerate}

Among the active joints, direct-driven modules (RJSIIT-17-RevB2 and RJSIIT-17-RevB5) are used for shoulder movements (Joints 1 and 2) to enhance robustness and provide high output torque during dynamic limb motion. 
The cable-driven joints are equipped with SEAs (AK80-64), providing benefits such as lightweight transmission and improved shock absorption. Potentiometers and encoders are installed on the joint side (3590S-2-104L for internal and external rotation joints and QY2204-SSI for the elbow joint) to ensure precise position sensing, as shown in Fig.~\ref{device}.
To enable accurate torque assistance, a full-joint torque sensing capability is achieved by integrating torque sensors (TK17-191151) on all cable-driven joints, ensuring precise execution of torque command during assistance.

\begin{table}[h!]
\centering
\caption{Functions of Exoskeleton Joints}
\begin{tabular}{c|l}
\hline
\textbf{Joint} & \textbf{Function} \\
\hline
Joint 0 & Shoulder eccentric movement \\
Joint 1 & Shoulder abduction/adduction \\
Joint 2 & Shoulder flexion/extension \\
Joint 3 & Upper-arm internal and external rotation \\
Joint 4 & Elbow flexion/extension \\
Joint 5 & Forearm internal and external rotation \\
\hline
\end{tabular}
\label{Joints_func}
\end{table}

% A cable-driven actuation system equipped with SEAs is employed for three active joints in the exoskeleton, providing benefits such as lightweight transmission and improved shock absorption. As illustrated in Fig.~\ref{device}, each cable-driven joint integrates a harmonic-drive servo motor, an encoder, and a pair of potentiometers.
% The actuation system relies on a linkage mechanism involving cables and springs, enabling force measurement through the compression of integrated springs, as captured by the potentiometers.

Considering the cable-driven mechanism and SEA, the dynamic model of the exoskeleton is expressed as follows:
\begin{align}
\bm M(\bm q)\ddot{\bm q} + \bm C(\dot{\bm q}, \bm q)\dot{\bm q} + \bm g(\bm q) &= \bm S_1 \bm u + \bm S_2^{\mathsf{T}} \bm K (\bm\theta- \bm S_2 \bm q)+ \notag \\
&\quad \bm \tau_{e} +\bm S_2^{\mathsf{T}} \bm \tau_{f},\label{dynRobot}\\
\bm B\ddot{\bm \theta} + \bm K (\bm\theta-\bm S_2 \bm q) &= \bm S_2\bm u,\label{dynSEA}
\end{align}
{where $\bm\tau_f$ represents the structured disturbance caused by friction in the cable-driven mechanism. The notations for the dynamics are detailed in Table~\ref{dynamic_table}.} 
%Specifically, $\bm\tau_o\hspace{-0.05cm}=\hspace{-0.05cm}\bm K(\bm\theta\hspace{-0.05cm}-\hspace{-0.05cm}\bm q)$ is the output torque of the SEA, connecting both the robot-joint subsystem (\ref{dynRobot}) and the motor-side subsystem (\ref{dynSEA}), and $\bm\tau_f$ represents unknown disturbance (e.g., friction) from the cable-driven mechanism. 
Because the direct-driven and cable-driven joints are mechanically decoupled, selection matrices are defined as follows:
\begin{align}
\bm{S}_1 &= \text{diag}(\underbrace{1, \ldots, 1}_{n-n_c}, \underbrace{0, \ldots, 0}_{n_c}) \in \mathbb{R}^{n \times n}, \\
\bm{S}_2 &= 
\begin{bmatrix}
\bm{0}_{n_c \times (n - n_c)} & \bm{I}_{n_c}
\end{bmatrix}
\in \mathbb{R}^{n_c \times n}.
\end{align}

\begin{table}[h]

\caption{Dynamic parameters}
\centering
\begin{tabular}{c|l}
\hline
% \textbf{Symbol} & \textbf{Brief Introduction} \\ \hline
$\bm M(\bm q)\in\mathbb{R}^{n\times n}$ & Inertia matrix of robot \\ \hline
$\bm C(\dot{\bm q}, \bm q)\in\mathbb{R}^{n\times n}$ & Matrix related to centripetal and Coriolis forces \\ \hline
$\bm g(\bm q)\in\mathbb{R}^n$ & Vector related to gravity \\ \hline
$\bm K\in\mathbb{R}^{n_c\times n_c}$ & Stiffness matrix \\ \hline
$\bm q\in\mathbb{R}^n$ & Robot joint angles\\ \hline
$\bm B\in\mathbb{R}^{n_c\times n_c}$ & Inertia matrix of motor \\ \hline
$\bm \theta\in\mathbb{R}^{n_c}$ & Motor rotation angles\\ \hline
$\bm\tau_e\in\mathbb{R}^n$ & Interaction torque with human participant \\ \hline
$\bm\tau_f\in\mathbb{R}^{n_c}$ & Disturbance torque \\ \hline
$\bm u\in\mathbb{R}^n$ & Control input exerted on robot joints \\ \hline
\end{tabular}
\begin{tablenotes}
\small
\item[1] $n$ is the number of DoFs.
\item[2] $n_c$ is the number of cable-driven DoFs.
\end{tablenotes}
\label{dynamic_table}
\end{table}

\begin{figure}[!t]
    \centering
    \includegraphics[width=0.9\linewidth]
    {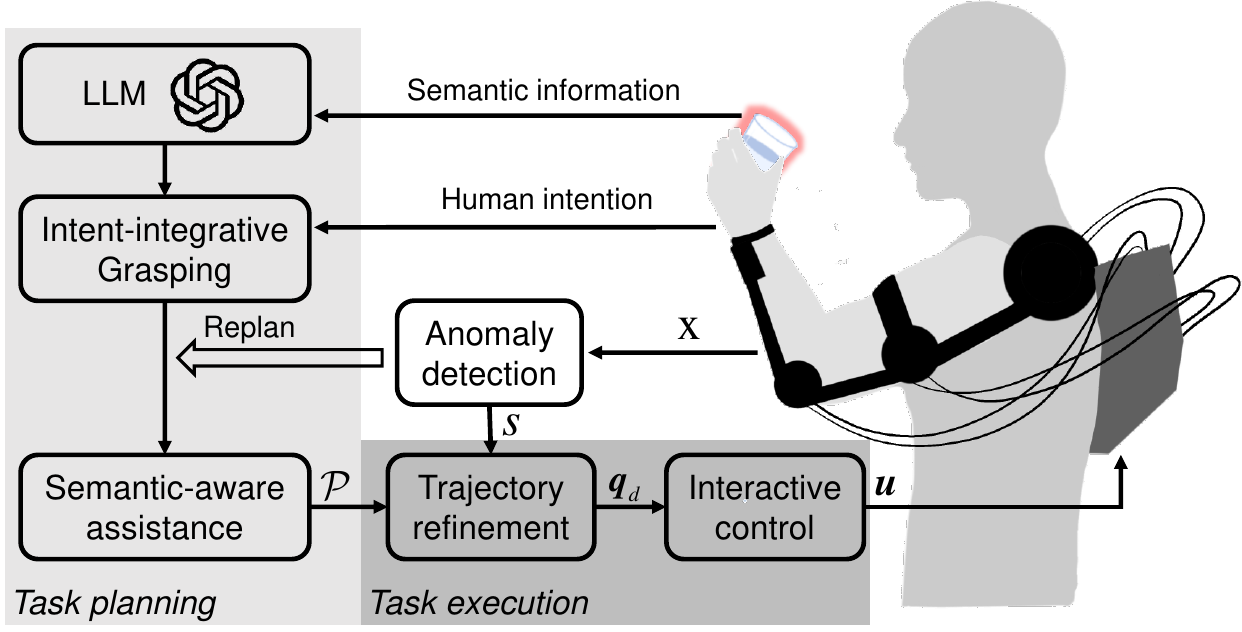}
    \caption{Block diagram of the proposed method: Semantic information is extracted by the LLM to enable task planning. Human intention intervenes during the grasping phase and subsequently receives assistance from the exoskeleton. The anomaly score $s$, computed from sensory feedback $\mathbf{x}$, guides online trajectory refinement and serves as a trigger for replanning, enhancing robustness to external anomalies. The reliable interactive control is ensured by an impedance controller designed to account for the cable-driven SEA mechanism. }
    \label{overall_frame}
    \vspace{-0.5cm}
\end{figure}

The proposed semantic-aware framework for upper-limb assistance is illustrated in Fig.~\ref{overall_frame}. Implementing a comprehensive assistance strategy that ensures both effectiveness and user-friendliness presents several challenges due to the following factors:
\begin{enumerate}
    \item [-] Unlike manipulators, where LLMs are implemented in isolation~\cite{brohan2023can}, exoskeletons operate in collaboration with human users. This requires the exoskeleton to identify the wearer’s motion intentions before initiating assistance. Furthermore, the system must incorporate a replanning mechanism to resolve any conflicts between the user's intention and the assistance being provided.

    \item [-] Because the exoskeleton augments the user’s movements, task planning must incorporate human factors. Semantic information from the task environment must be accurately interpreted and used to reason through solutions that reflect the user's needs and perspectives.

    \item [-] During movement execution, the exoskeleton must account for potential risks associated with limb movement. Real-time regulation of human-robot interaction is essential to ensure safety throughout the assistance process.
\end{enumerate}
This paper addresses these challenges through the proposed semantic-aware framework, which consists of two main phases: task planning and task execution. The task planning phase focuses on providing semantic-aware assistance by interpreting and using semantic information from the environment, ensuring that the planned tasks align with human intentions and contextual requirements. The task execution phase complements this by adjusting configurable parameters in the control scheme, enabling real-time regulation of human-robot interaction.

\section{Task Planning}
In tasks involving object manipulation and handling, the wearer requires additional support from the exoskeleton when interacting with objects. 
{In other words, the transparent mode and the LLM-involved assistance are functionally decoupled. The exoskeleton remains transparent during the grasping phase, allowing the wearer to perform the grasp autonomously, thereby minimizing potential motion interference and discomfort. Assistance is provided only during the subsequent object transportation phase.}
Based on this, we divide the planning of such manipulation tasks into two components: \textit{Intent-Integrative Grasping} and \textit{Semantic-Aware Assistance}. The former uses human intention to control the exoskeleton for flexible and reliable grasping, whereas the latter selects effective assistance parameter configurations based on task semantics, ensuring safety is not compromised.\\

\noindent\textbf{Intent-Integrative Grasping}: 
For tasks that require a grasping stage, this framework delegates the sub-task, i.e., grasping, to the wearer, allowing human intention to control the exoskeleton in selecting the target and achieving stable grasping.
{This approach effectively avoids failures associated with predefined fixed grasp positions.}
If these actions were handled by the exoskeleton, additional sensors would be required to enable full environmental observation and understanding. This could result in less comfortable assistance during free limb movement and might compromise the system's compactness. However, these actions are inherently easy for humans to perform and do not require additional assistance. To enable the wearer to effortlessly maneuver the exoskeleton during grasping, we employ a transparent mode where the controller is designed as follows:
\begin{align}
    \bm u = &(\bm M(\bm q)+\bar{\bm B})\ddot{\bm q}_0+\bm C(\dot{\bm q}, \bm q)\dot{\bm q}\notag\\
    &+\bm g(\bm q)-\bm S_2^{\mathsf{T}}\hat{\bm\tau}_f-  {\bm\tau}_e + \bm u_f,\\
    \ddot{\bm q}_0 = &\frac{1}{\gamma_0}\bm (\bm M(\bm q)+\bar{\bm B})^{-1}{\bm\tau}_e,
\end{align}
where $\gamma_0$ is a scale factor; $\bm u_f\triangleq-\bm S_2^{\mathsf{T}}\bm K_v(\dot{\bm\theta}-\bm S_2 \dot{\bm q})$, with $\bm K_v$ defined as a diagonal and positive-definite matrix, is the control term that decouples the fast and slow system in the SEA based on the singular perturbation~\cite{shu2023twostage}; and $\bar{\bm B} = \bm S_2^{\mathsf{T}}\bm B \bm S_2$ is the projected motor inertia matrix in the joint side~\cite{chen2024upper}.
{In such a mode, the exoskeleton achieves a backdrivability torque of less than 0.7 N$\cdot$m at 10 $^{\circ}$/s, enabling it to compliantly follow the wearer's upper-limb movements during the grasping phase.}

\begin{figure}[!t]
    \centering
    \includegraphics[width=0.75\linewidth ]{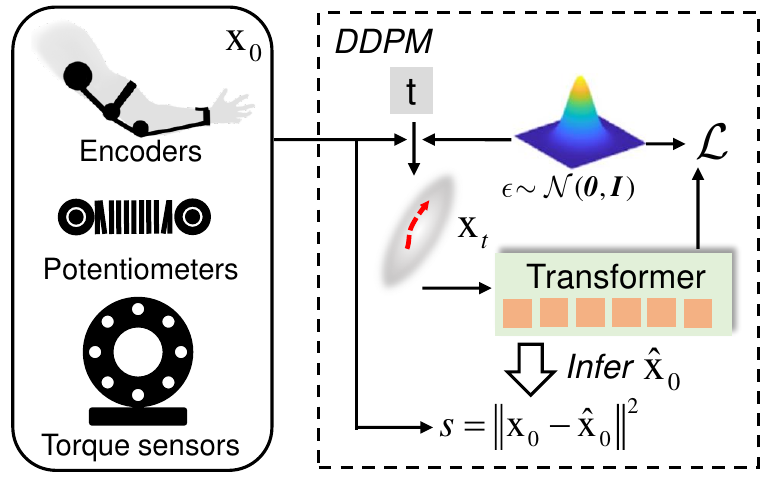}
    \caption{Illustration of the anomaly detector: The multi-modal sensory feedback, which includes position, velocity, and interaction torque information, is represented by $\mathbf{x}$.
    The diffusion model learns the data distribution under normal interaction states and reconstructs sensory feedback to determine the anomaly level in real time.}
    \label{ano_net}
    \vspace{-0.5cm}
\end{figure}

Intention is incorporated not only in selecting the grasping target but also continuously as human feedback throughout the various stages of assistance. To this end, we introduce an anomaly detection network based on a diffusion model to assess the alignment between the wearer and the exoskeleton, as illustrated in Fig.~\ref{ano_net}.
{The detector can be trained in an unsupervised manner and has been validated across different upper-limb movements~\cite{chen2024upper}.}
The detector is trained on sensory feedback $\bm{\mathbf{x}}^{(i)}\in\hspace{-0.05cm}\Re^{L_{s}N_c}$ collected while the wearer performs arbitrary upper-limb movements in transparent mode, where the superscript $i$ denotes the time step, $L_s$ is the sliding-window length, and $N_c$ is the number of sensing channels. The diffusion model operates on a sequence $(\bm{\mathbf{x}}_{T}, \bm{\mathbf{x}}_{T-1}, \cdots, \bm{\mathbf{x}}_0)$ over a duration $T$, defining both the forward diffusion and the reverse denoising processes as follows:

% Intention is integrated not only in the selection of a grasping target but also continuously throughout various stages of assistance as human feedback. To achieve this, we propose an anomaly detection network based on a diffusion model to evaluate the alignment between the wearer and exoskeleton. The underlying principle of this neural network is illustrated in Fig.~\ref{ano_net}.
% The anomaly detector is trained using sensory feedback $\bm{\mathbf{x}}^{(i)}\in\hspace{-0.05cm}\Re^{L_{s}N_c}$ collected when the user wears the exoskeleton and performs arbitrary upper-limb movements in transparent mode.
% Here, the superscript $i$ denotes the time step, $L_s$ is the width of the sliding window, and $N_c$ is the number of data channels.
% To construct the diffusion model, a sequence $(\bm{\mathbf{x}}_{T}, \bm{\mathbf{x}}_{T-1}, \cdots, \bm{\mathbf{x}}_0)$ of duration $T$ is introduced, defining both the diffusion and reverse diffusion processes as follows:
\begin{align}
    q(\bm{\mathbf{x}}_{t}|\bm{\mathbf{x}}_{t-1}) &= \mathcal{N}(\bm{\mathbf{x}}_{t};\sqrt{1-\beta_{t}}\bm{\mathbf{x}}_{t-1},\beta_{t}\bm I),\label{diffusion_process}\\
    q_{\Psi}(\bm{\mathbf{x}}_{t-1}|\bm{\mathbf{x}}_{t}) &= \mathcal{N}(\bm{\mathbf{x}}_{t-1};\bm\mu_{\Psi}(\bm{\mathbf{x}}_{t}),\Tilde{\beta}_{t}\bm I),\label{reverse_process_AD}
\end{align}
where $(\beta_1,\beta_2,\cdots,\beta_{T})$ represents the variance schedule that regulates the injected noise, and $\Tilde{\beta}_{t}=\frac{1-\bar{\alpha}_{t-1}}{1-\bar{\alpha}_t}\beta_t$ is the adjusted variance with the definitions $\alpha_t = 1-\beta_t$ and $\bar{\alpha}_t = \prod_{i=1}^{t}\alpha_i$. The learnable parameters are denoted as $\Psi$. 
By adopting the variational lower bound as the training loss, we transform the optimization process into the noise space $\bm \epsilon \thicksim \mathcal{N}(\bm 0, \bm I)$ using parameterization techniques~\cite{ho2020denoising}:
\begin{align}
    \mathcal{L}(\Psi) = \mathbb{E}_{t,\bm{\mathbf{x}}_0,\bm \epsilon}[\|\bm \epsilon - \bm \epsilon_{\Psi}(\bm{\mathbf{x}}_t)\|^2].
\end{align}

Once the reverse process is learned, the sensory feedback can be reconstructed as $\hat{\mathbf{x}}_0$ through denoising:
\begin{align}
    \hat{\mathbf{x}}_\nu =& \mathbf{x}_0 \sqrt{\bar{\alpha}_\nu} + \bm \epsilon_1\sqrt{1-\bar{\alpha}_\nu},\\
    \hat{\mathbf{x}}_{t-1} =& \frac{1}{\sqrt{\alpha_t}}(\hat{\mathbf{x}}_{t}-\frac{1-\alpha_t}{\sqrt{1-\bar{\alpha}_t}}\bm \epsilon_{\Psi}(\hat{\mathbf{x}}_t))+\Tilde{\beta}_{t}\bm \epsilon_2,
\end{align}
where $\bm \epsilon_1,\bm \epsilon_2\thicksim \mathcal{N}(\bm 0, \bm I)$ and $\nu \in [1, T]$ is the constant parameter that controls the injected noise.
The reconstruction result indicates the similarity between the current human-robot interaction state and training data. 
The corresponding anomaly score is defined as~\cite{zhang2023multi}
\begin{align}
s = {||\mathbf{x}_0- \hat{\mathbf{x}}_0||}^{2}=f(\bm q, \dot{\bm q}, \bm \theta, \dot{\bm \theta},\bm \tau_e),
\label{AD_function}
\end{align}
where $f(\cdot)$ represents the anomaly score calculation process, and the function inputs represent specific sensory feedback signals.
{Since the anomaly score continuously monitors the quality of human-robot interaction in real time, it serves as a critical trigger for adaptive behavior.
To respond to potential conflicts between the exoskeleton's actions and the user's intentions, the proposed framework incorporates a replanning mechanism.} 
When the anomaly score exceeds a predefined threshold, indicating the conflict between exoskeleton and human intention, the system initiates a replanning process, as detailed in Algorithm~\ref{LLM_ALG}.\\

\begin{figure*}[!t]
    \centering
    \includegraphics[width=0.8\linewidth ]{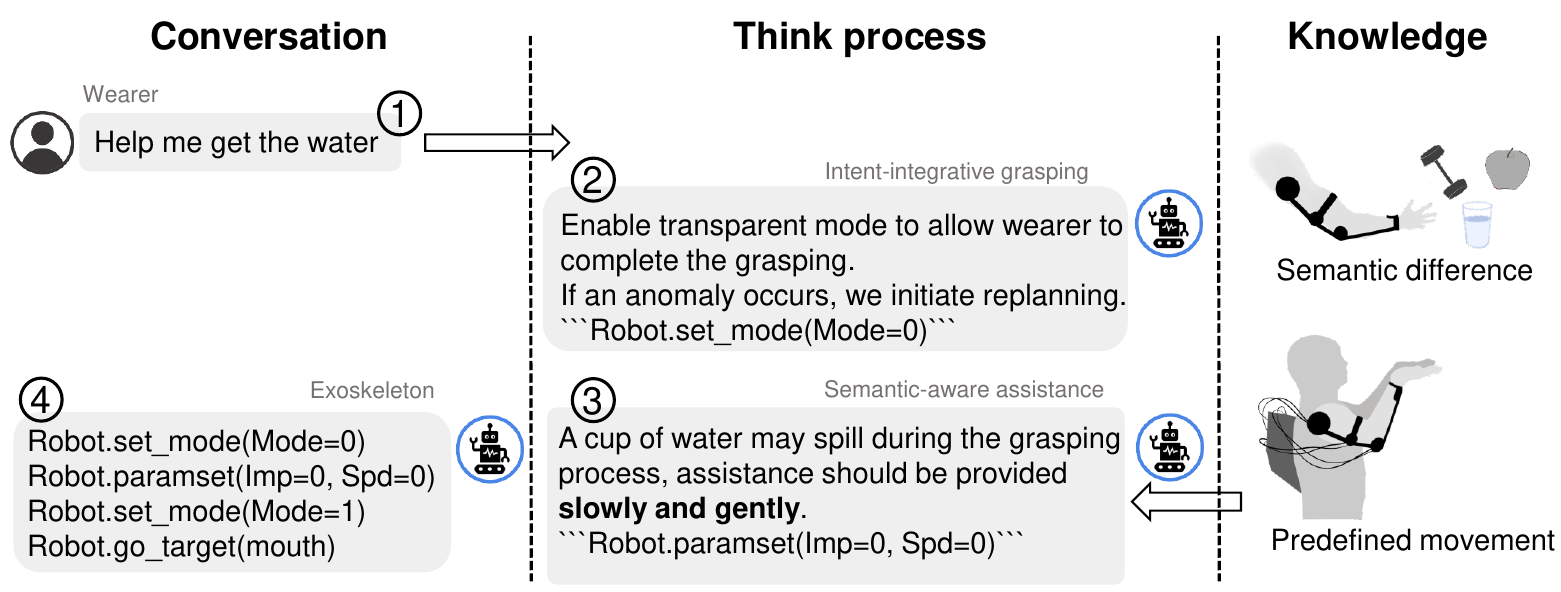}
    \caption{{The semantic-aware assistance workflow begins with the LLM activating transparent mode for user-initiated actions like grasping.
    In this study, the exoskeleton did not assist grasping, allowing the wearer to directly manage variations in object shape and size with their own hand. It then extracts key semantic information, such as potential danger, need for gentleness, or predefined motions, to adapt the exoskeleton’s behavior. Based on this, three binary parameters are configured: control mode (0 for transparent, 1 for impedance), assistance intensity (0 for low, 1 for high stiffness and damping), and movement speed (0 for low, 1 for high speed).}}
    \label{chat_flow}
    \vspace{-0.5cm}
\end{figure*}

\begin{algorithm}[t]
\caption{Task Planning with Anomaly Detection}
\begin{algorithmic}[1]
\REQUIRE Description $l_{\Pi},l_{T}$, sensory feedback $\mathbf{x}$
\ENSURE Plan $\mathcal{P}$
\STATE $n=0,\pi=\varnothing,\mathcal{P}=\varnothing$
\WHILE{$l_{\pi_{n-1}} \neq \textit{"done"}$}
    \STATE $\mathcal{S}_1,\mathcal{S}_2=\varnothing$
    \STATE calculate anomaly score $s=f(\mathbf{x})$
    \FOR{$\pi \in \Pi$ and $l_{\pi} \in l_{\Pi}$}  
        \STATE $p_{\pi}=p(l_{\pi}|l_p,l_T)$
        \STATE $\mathcal{S}_1=\mathcal{S}_1\bigcup p_{\pi}$
    \ENDFOR
    \STATE $\pi=\arg \max_{\pi \in \Pi}\mathcal{S}_1$
    \FOR{$\pi_c \in \pi$ and $l_{\pi_c} \in l_{\pi}$}
        \STATE $p_{\pi_c}=p(l_{\pi_c}|l_p,l_T,l_{\pi})$
        \STATE $\mathcal{S}_2=\mathcal{S}_2\bigcup p_{\pi_c}$
    \ENDFOR
    \STATE $\pi_{n}=\arg \max_{\pi_c \in \pi}\mathcal{S}_2$
    \STATE $\mathcal{P}=\mathcal{P}\bigcup {\pi_c}$
    \STATE $n=n+1$
    \IF{$s\geq \bar{s}$}
    \STATE $\mathcal{P}=\varnothing$
    \STATE $n=0$
    \ENDIF
\ENDWHILE
\end{algorithmic}
\label{LLM_ALG}
\end{algorithm}

\noindent\textbf{Semantic-Aware Assistance}: 
Although LLMs can facilitate robot control through language, the semantic information embedded in instructions is often overlooked, limiting their effectiveness in task planning and comprehensive assistance.
{To address this, we query LLMs by text input to generate semantic-aware assistance that can be directly executed, as depicted in Fig.~\ref{chat_flow}.}
{By leveraging semantic-aware assistance, inappropriate assistive strategies can be avoided, thereby enhancing the overall assistance performance.}
The knowledge acquired from the training corpora enables the model to recognize target semantic attributes in grasping tasks or to interpret and execute other tasks via predefined movements, such as preparing to catch an object. We define a subtask set $\Pi$ comprising predefined high-level movements with configurable parameters, where each subtask $\pi$ is linked to a corresponding description $l_{\pi}$. Given a task description $l_T$, the LLM selects an optimal subtask sequence using the following strategy:
\begin{align}
    \pi = \arg \max_{\pi \in \Pi}p(l_{\pi}|l_p,l_T),
\end{align}
where $l_p$ serves as a prompt containing descriptions of available functions and examples of successful task planning. To enhance reasoning capabilities and ensure valid task planning, the chain-of-thought (CoT) approach~\cite{wei2022chain} is incorporated into the planning process.

For configurable subtask $\pi_c\in\pi$, the LLM extracts semantic information from the task description to appropriately adjust the task parameters. For instance, if the task involves carrying a heavy object, a higher impedance parameter should be applied; meanwhile, when handling fragile items, a lower speed constraint should be enforced. The optimal subtask configuration is determined as follows:
% In this paper, the configurable parameters are shown in Table~\ref{configurable_para}
\begin{align}
    \pi_c = \arg \max_{\pi_c \in \pi}p(l_{\pi_c}|l_p,l_T,l_{\pi}),
\end{align}
where $l_{\pi_c}$ denotes the description linked to the subtask configuration $\pi_c$, and the complete task planning process is presented in Algorithm~\ref{LLM_ALG}. This approach leverages semantic information to optimize task execution through LLM-based subtask selection and configuration, and, when integrated with an anomaly-driven replanning mechanism, provides comprehensive and safe assistance that remains seamlessly aligned with the wearer’s intentions.\\

% \begin{table}[h]
% \centering
% \caption{Configurable parameters in subtask}
% \begin{tblr}{
%   hline{1,4} = {-}{0.08em},
%   hline{2} = {-}{},
% }
% Mode & $\bm K_d,\bm C_d$ & $\mathcal{X}$ \\
% Transparent    & Low       & Low   \\
% Impedance    & High      & High  
% \end{tblr}
% \label{configurable_para}
% \end{table}

\section{Task Execution}
Once the task is planned by the LLM, the exoskeleton executes each subtask in sequence while adapting its configuration based on the extracted semantic information, achieved through two integrated mechanisms: \textit{online trajectory refinement}, which generates the assistance trajectory for each subtask, and \textit{interactive control}, which regulates human–robot interaction within a desired impedance model, allowing speed and impedance parameters to be adjusted throughout execution.\\

\noindent\textbf{Online Trajectory Refinement}: For subtasks that require active assistance from the exoskeleton, a feasible trajectory is generated and refined while prioritizing human comfort.
Assuming that assistance is provided in free space, where no obstacles are present along the assistive path, the reference trajectory $\bm q_r$ is constructed using quintic spline interpolation between the current and target poses, with a tunable duration $t_f$. 
To ensure that the refined trajectory adheres to dynamic constraints and promotes comfort, we define the state vector $\bm y_d = [\bm q_d^{\mathsf{T}},\dot{\bm q}_d^{\mathsf{T}}]^{\mathsf{T}}$ and formulate the trajectory refinement process as a quadratic programming problem:
\begin{align}
\label{opt_all}
\min_{\bm y_d, \bm u_d, s} &\sum_{i=t}^{t+N}\left[\|\bm q_{d}^{(i)} - \bm q_{r}^{(i)}\|_{\bm Q}^2 + \|\bm u_{d}^{(i)}\|_{\bm R}^2 + (s^{(i)})^2\right],\\
 s.t.\quad &{\bm y}_d^{(t+1)} = \begin{bmatrix}\bm I & \bm I\Delta t\\\bm 0 & \bm I\end{bmatrix}\bm y_d^{(t)} + \begin{bmatrix}\bm 0\\\bm I\Delta t\end{bmatrix}\bm u_d^{(t)}\notag\\
 &s^{(t+1)}=s^{(t)} + \begin{bmatrix}\bm 0\\-(\frac{\partial f}{\partial \bm \tau_e})^{\mathsf{T}}\bm K_{d}w\Delta t\end{bmatrix}\bm x_d^{(t)}\notag\\
&\qquad\qquad-(\frac{\partial f}{\partial \bm \tau_e})^{\mathsf{T}}\bm C_{d}\bm u_d^{(t)}w\Delta t\notag\\
&\bm y_d \in \mathcal{Y}, \bm u_d \in \mathcal{U} \notag
\end{align}
where $\bm q_d$ is the refined desired trajectory from the reference $\bm q_r$, optimized over horizon $N$ with step $\Delta t$ under symmetric positive-definite weights $\bm Q$ and $\bm R$, subject to trajectory and acceleration constraints $(\mathcal{Y},\mathcal{U})$, where the tunable speed limit $ \dot{\bm q}_d \leq \bar{\dot{\bm q}}_d$ directly modulates the assistive trajectory during execution.\\

% where $\bm q_d$ represents the refined desired trajectory derived from the reference trajectory $\bm q_r$, $N$ denotes the horizon, and $\Delta t$ is the discrete time step. The matrices $\bm Q$ and $\bm R$ are symmetric positive-definite weighting matrices, $\bm u_d$ corresponds to the acceleration of the desired trajectory, and $\mathcal{Y}$ and $\mathcal{U}$ define the trajectory space and acceleration constraints, respectively.
% Among these, the speed constraint $ \dot{\bm q}_d\leq \bar{\dot{\bm q}}_d$ is a tunable parameter during task execution, which directly modulates the assistive trajectory after task planning.

\noindent\textbf{Interactive Control}: 
Because assistance is delivered through the interaction between the exoskeleton and wearer, an impedance model~\cite{Zhang2023MultiModalLA} is defined to regulate the human-robot interaction:
\begin{eqnarray}
&\bm C_d(\dot{\bm q}-\dot{\bm q}_d)+\bm K_d(\bm q-\bm q_d)=\frac{1}{w}\bm\tau_e,\label{IC_model}
\end{eqnarray}
where $\bm C_{d}$ and $\bm K_{d}$ are the impedance parameters, which are adjustable through $w$, a scaling factor that modulates the impedance level in response to varying semantic conditions.
{To implement this model during assistance and compensate for disturbances, a singular perturbation-based impedance controller is used \cite{chen2024upper} along with polynomial fitting techniques for addressing cable-induced hysteresis.}
By substituting the desired impedance model (\ref{IC_model}) into (\ref{AD_function}), the anomaly score transition in (\ref{opt_all}) is obtained, enabling trajectory refinement to minimize anomalies during assistance, where the configurable parameters $\mathcal{C}=(t_f,\bar{\dot{\bm q}}_d,w)$ are tuned according to the target’s semantic attributes to adapt speed and assistance intensity for safe and effective human–robot interaction.\\

\section{Experiments}
\begin{figure}[!t]
    \centering    \includegraphics[width=0.6\linewidth]{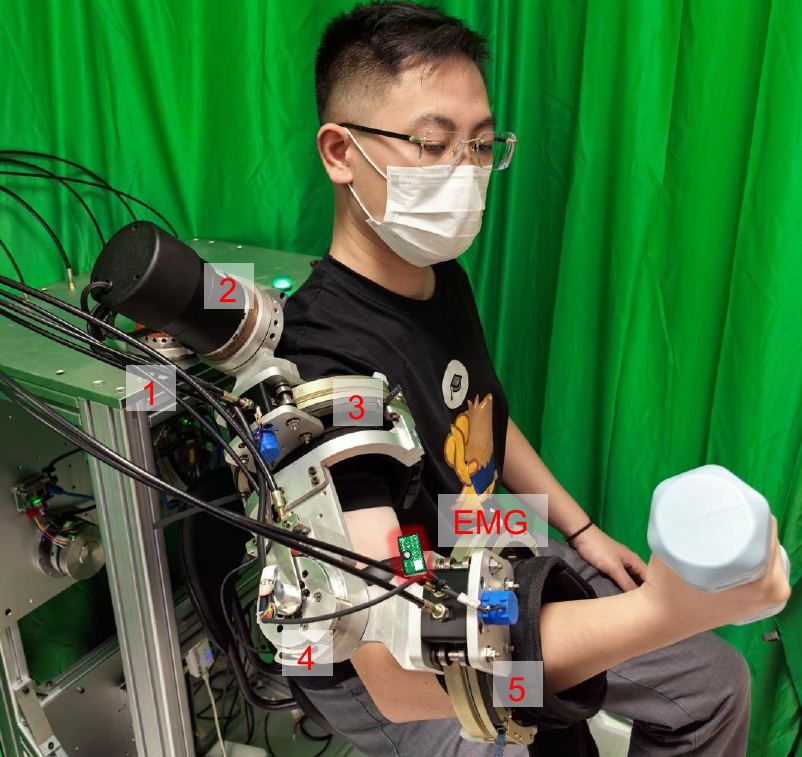}
    \caption{{The experimental setup consists of an upper-limb exoskeleton featuring directly driven joints (Joints 1 and 2) and cable-driven SEA joints (Joints 3, 4, and 5). An electromyography sensor is employed to detect successful grasp events based on muscle activity.}}
    \label{real_device}
    \vspace{-0.5cm}
\end{figure}

% \subsection{Disturbance Compensation}
% In order for the exoskeleton to deliver AAN support to the patient, it is essential for the robot to acquire authentic motion patterns from a healthy wearer. 
Experiments with the developed upper-limb exoskeleton were conducted to validate the proposed framework for delivering semantic-aware assistance. The experimental setup is illustrated in Fig.~\ref{real_device}. Because of the absence of a rigid connection between the wearer and exoskeleton, assistance for internal and external forearm rotation was limited. Thus, only Joints 1, 2, 3, and 4, corresponding to the shoulder and elbow joints, were activated to provide assistance.
A separate PC interfaced with the exoskeleton through the controller area network bus to communicate with the LLMs. 
{To segment the grasping phase from subsequent assistance, an electromyography sensor (Ws450, Biometrics Ltd.) was attached to the wearer's arm and used as a trigger mechanism, enabling the wearer to confirm a stable grasp through intentional muscle contractions.}

\begin{table}[!h]
\vspace{-0.5cm}
\centering
\caption{Robot command setting}
\begin{tblr}{
  cell{2}{1} = {r=2}{},
  cell{2}{2} = {r=2}{},
  cell{4}{1} = {r=4}{},
  cell{4}{2} = {r=2}{},
  cell{6}{2} = {r=2}{},
  vlines,
  hline{1-2,4,8-10} = {-}{},
  hline{3,5,7} = {3}{},
  hline{6} = {2-3}{},
}
Function   & Input   & Description                 \\
\texttt{set\_mode}  & Mode    & 0: transparent mode          \\
           &         & 1: impedance mode           \\
\texttt{paramset}   & Imp     & 0: $w=2$                      \\
           &         & 1: $w=0.5$                    \\
           & Spd     & 0: $(t_f,\bar{\dot{\bm q}}_d)=(5s,30^{\circ}/s)$                \\
           &         & 1: $(t_f,\bar{\dot{\bm q}}_d)=(2s,70^{\circ}/s)$                \\
\texttt{go\_target} & Target~ & Move to target place        \\
\texttt{others}     & None    & Execute predefined movement 
\end{tblr}
\label{llm_func_set}
\end{table}

In the proposed semantic-aware framework, assistance was generated through a valid combination of multiple subtasks. These subtasks, implemented as executable Python functions with configurable parameters, are listed in Table~\ref{llm_func_set}.
The functions $\texttt{set\_mode}$ and $\texttt{paramset}$ are binary configurable and used to define different controller modes and adjust impedance and speed settings. In this study, the target position is predefined as a relative joint position with respect to the wearer, such as the mouth or chest. The predefined movements extend the proposed framework to support specific upper-limb tasks beyond grasping, such as preparing to catch an object or perform an arm swing. These movements can be preprogrammed as Python functions, serving as selectable tools during task planning and enabling generalization to other task scenarios.
{The anomaly detection model is trained using data obtained from two healthy male subjects, to improve its generalization capability across different users.}

The assistance process was divided into generation and delivery phases, which were separately evaluated to verify the proposed framework's ability to successfully perform task planning and provide comprehensive and effective assistance.\\  
\subsection{Assistance Generation}
\begin{figure}[!t]
    \vspace{-0.1cm}
    \centering
    \includegraphics[width=1\linewidth]{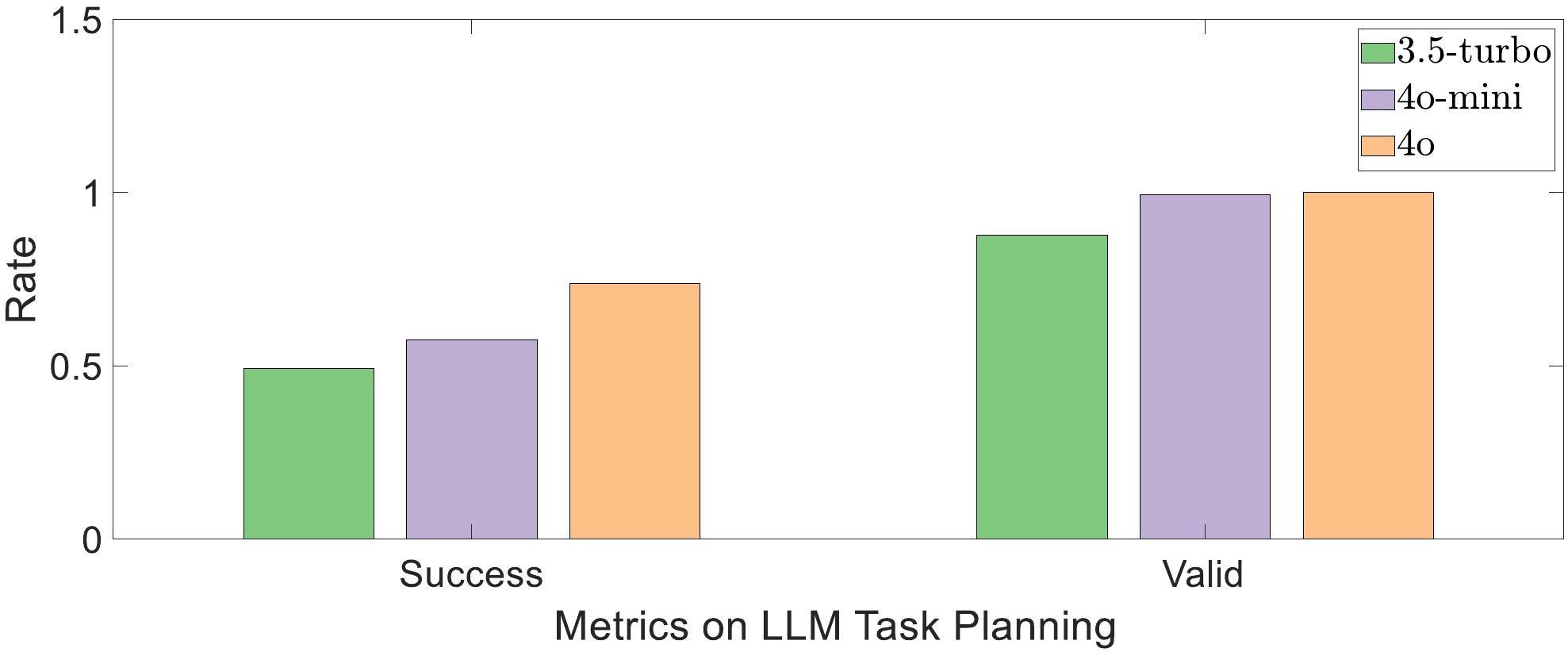}
    \caption{{Performance of different language models: GPT-4o surpasses other models in both success rate and adherence to predefined execution rules, where plans with infeasible steps are deemed invalid.}}
    \label{diff_model}
    \vspace{-0.5cm}
\end{figure}

%333333333333333333333333333333333333333333
Because task planning depends on the model's intrinsic knowledge to extract semantic information and configure assistance, we evaluated the effect of LLM intelligence on the method's performance.
{Specifically, we compared three models accessed via OpenAI’s commercial API, namely GPT-3.5 Turbo, GPT-4o Mini, and GPT-4o, using two evaluation metrics}: the success rate of extracting object semantic information and the validity of their output instructions. 
The first metric determines whether the model correctly identifies the need for high-impedance or slow-speed assistance to maintain operational stability. The second metric evaluates the model’s ability to perform task planning under predefined regulations, ensuring that output instructions are valid and do not include unrecognized tasks. 
{These metrics were assessed using 195 homecare-related items, which is categorized into three main types: (1) 50 food items, such as apples and carrots; (2) 85 everyday household objects that can be grasped and placed, including spoons and mugs, as well as potentially hazardous items such as scissors and cups containing hot liquids; and (3) 60 heavy objects, including dumbbells and suitcases.}
The results (Fig.~\ref{diff_model}) indicated that GPT-4o had the highest success rate and strictest rule adherence, making it the chosen base model.

\begin{figure}[!t]
    \centering
    \includegraphics[width=0.7\linewidth]{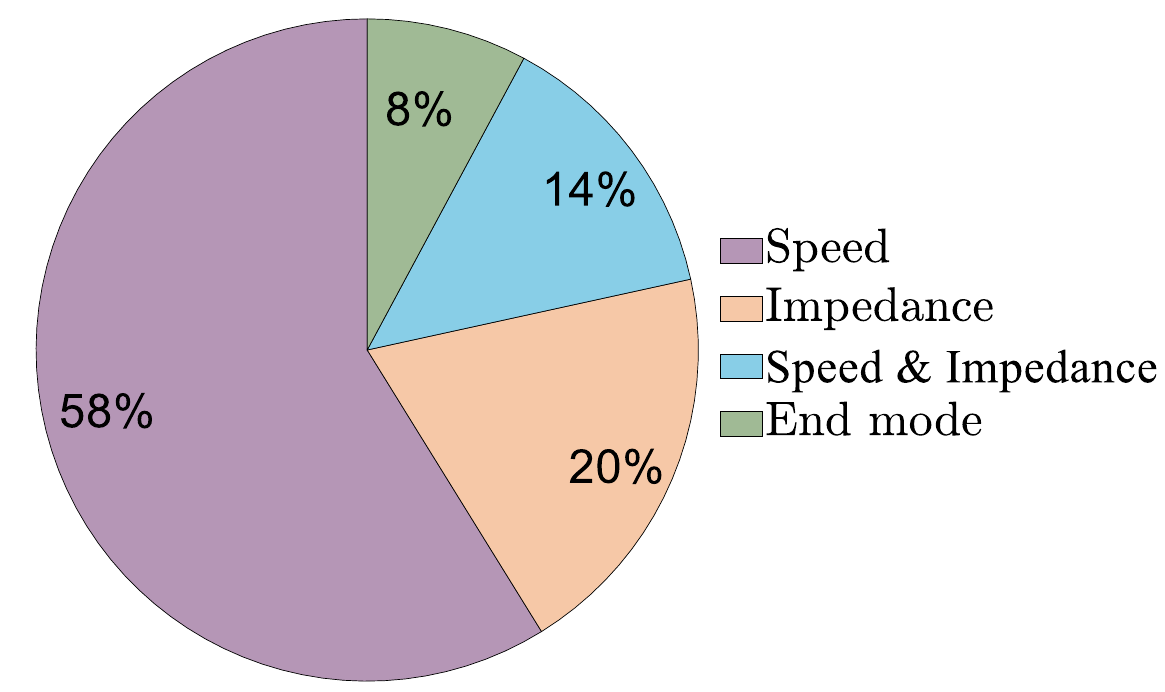}
    \caption{{Error breakdown of semantic-aware assistance with GPT-4o and CoT shows that lacking cognitive alignment leads to overly conservative speed settings.}}
    \label{error_pie}
    \vspace{-0.3cm}
\end{figure}

\begin{figure}[!t]
    \vspace{-0.1cm}
    \centering
    \includegraphics[width=1\linewidth]{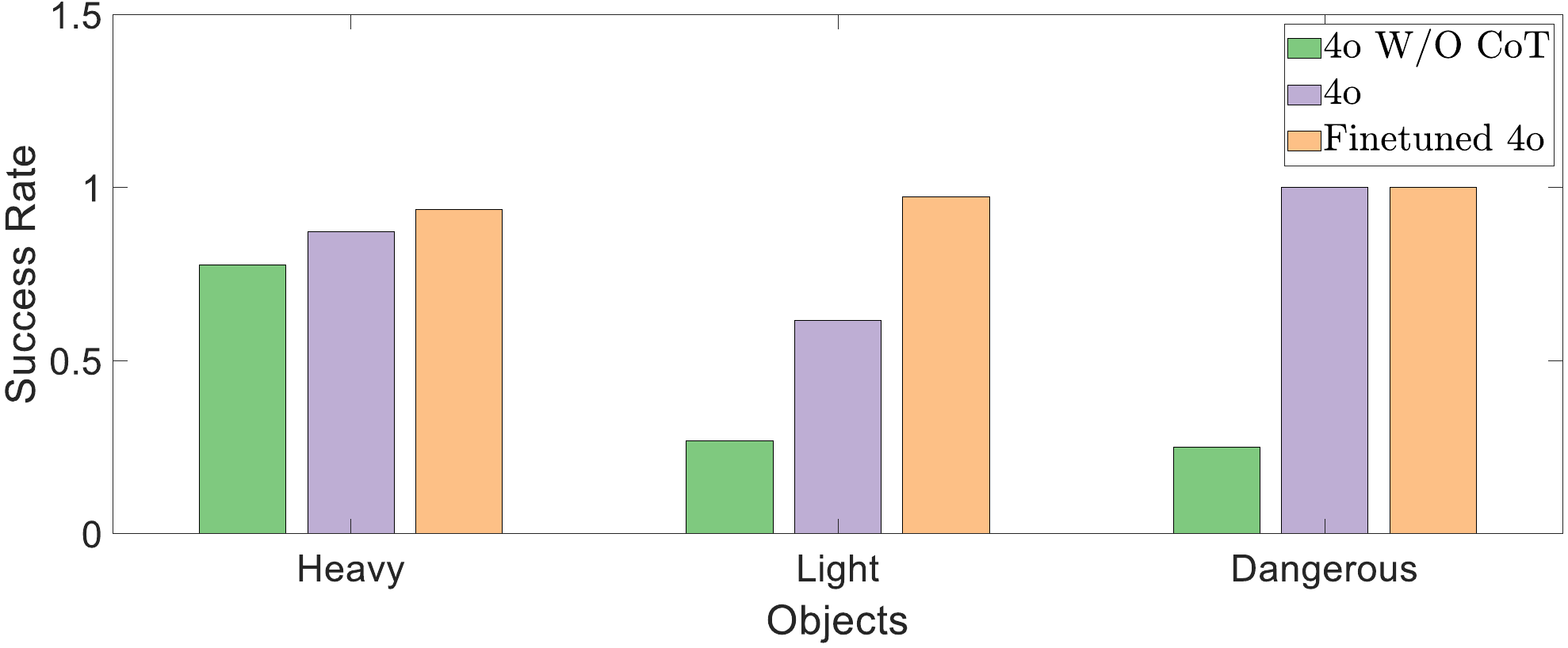}
    \caption{{Performance of LLM on semantic-aware assistance tasks under different configurations: Combining CoT with fine-tuned cognitive alignment yields the highest success rate across tasks.}}
    \label{diff_use}
    \vspace{-0.5cm}
\end{figure}

Fig.~\ref{error_pie} presents the error breakdown of item configurations when using GPT-4o with CoT reasoning, categorizing errors into impedance, speed, combined impedance and speed misconfigurations, and end mode errors. End mode errors occur when improper control modes are applied after handling heavy or hazardous objects, situations in which impedance control should have been maintained.
The primary source of error arises from conservative speed configurations. For instance, items such as cereal boxes, which are safe for high-speed handling, are instead operated at low speeds, reducing assistance efficiency. 
To mitigate these errors, CoT reasoning and fine-tuning techniques were applied to align LLM perception with that of the wearer. An ablation study, shown in Fig.~\ref{diff_use}, demonstrated that CoT reasoning improved accuracy by $92.1\%$ compared with direct task planning, particularly excelling in identifying dangerous items with $100\%$ accuracy, thereby preventing potential harm.
{In addition, a subset of 60 items was randomly selected from the full set of 195 homecare-related objects. These items were annotated with semantic labels and integrated into a dialogue-based dataset to fine-tune GPT-4o.
This fine-tuning process enhanced the model’s cognitive alignment with the wearer, leading to a $21\%$ improvement in distinguishing between heavy and light objects.}
These results confirm the successful cognitive alignment between LLMs and the wearer for appropriate parameter configurations.
{Furthermore, the framework’s generalizability was validated for both additional subjects and other predefined movement primitives, such as assisting with object holding and arm swinging, as demonstrated in the supplementary video.}

% An ablation study on the employed CoT technology was carried out to evaluate the accuracy of semantic extraction for different types of objects, as shown in Fig.~\ref{diff_use}. The CoT technology improves the $92.1\%$ compared to direct task planning, especially on the identification of the dangerous item, which may result in harm to the wearer, while the CoT technology helps to achieve the $100\%$ accuracy.
% Additionally, we also validated that the knowledge of LLMs could also be aligned with the wearer in the proposed pipeline. As the wearer identified the semantic information of 60 items, it was generated into a dialog to finetune the models to achieve alignment with the cognition of the wearer.
% The result shows that the accuracy in identifying heavy or light items increases $21\%$, indicating the successful alignment between LLMs and the wearer.
% Among the case of applying the 

\begin{figure}[!t]
\centering
\subfigure[]{
    \label{snap1}
    \includegraphics[width=1\linewidth]{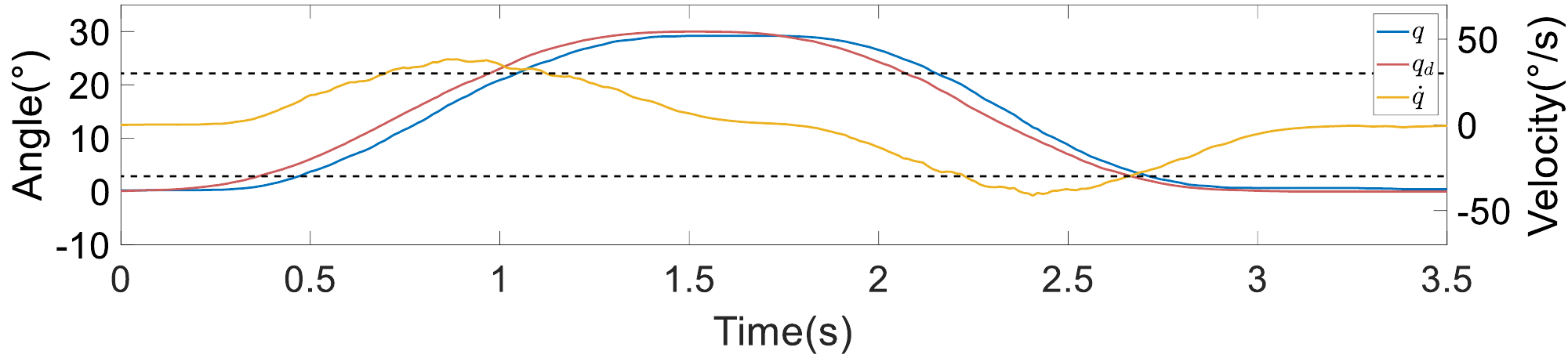}}
\subfigure[]{
    \label{snap2}
    \includegraphics[width=1\linewidth]{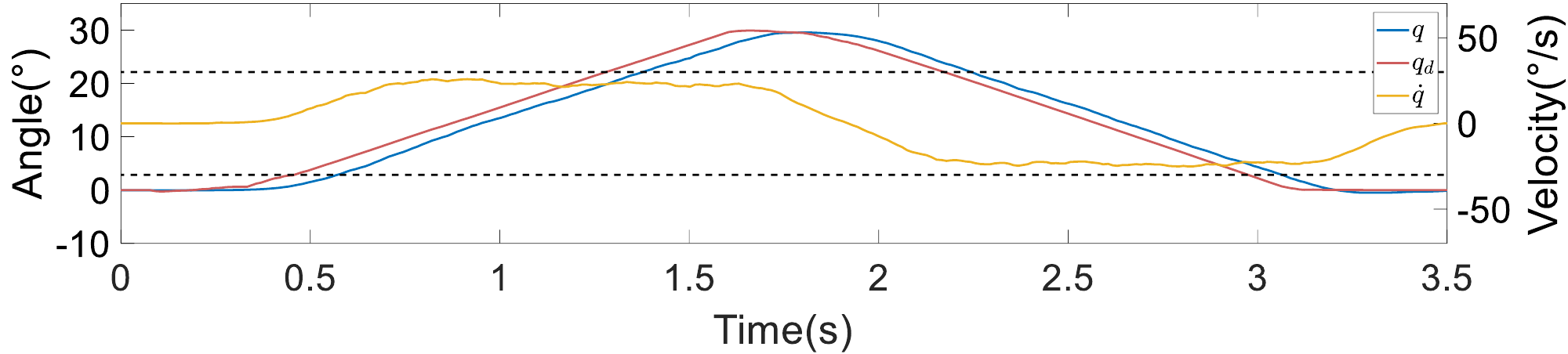}}
    \vspace{-0.5cm}
\caption{Tracking performance of the same reference trajectory under identical speed constraints: (a) without online refinement and (b) with online refinement. The black dashed line indicates the specified speed limit.}\label{refine_check}
\vspace{-0.3cm}
\end{figure}

\begin{figure}[!t]
\centering
\subfigure[]{
    \label{snap1}
    \includegraphics[width=0.31\linewidth]{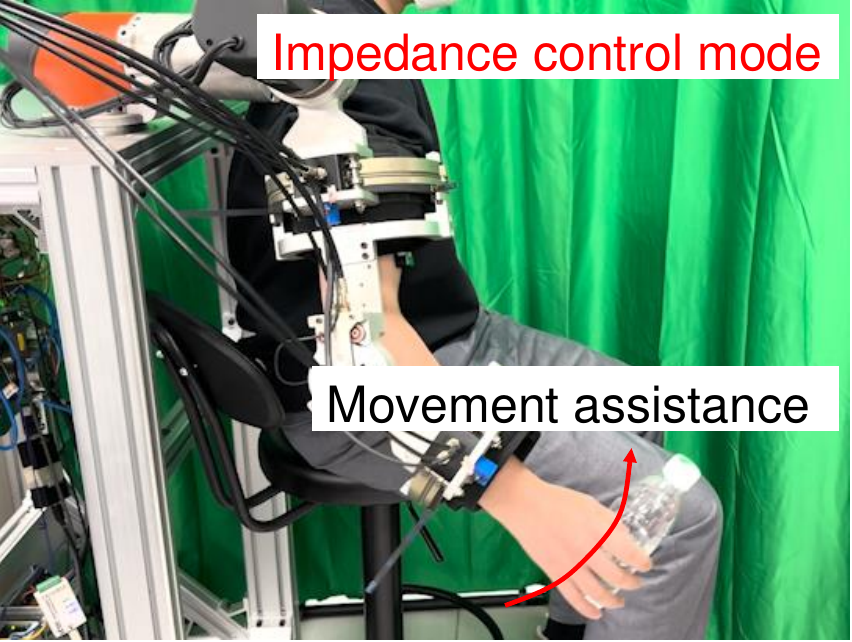}}
\subfigure[]{
    \label{snap2}
    \includegraphics[width=0.31\linewidth]{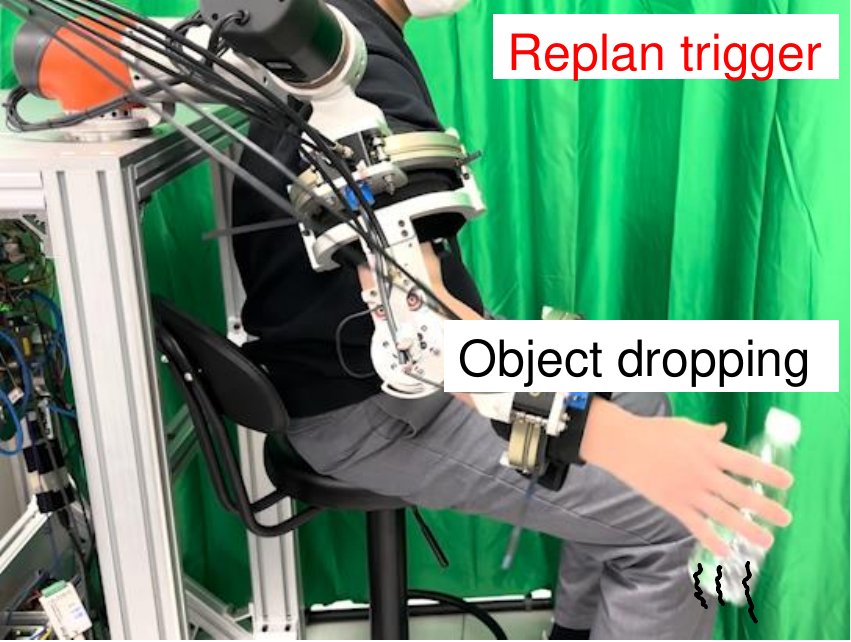}}
\subfigure[]{
    \label{snap3}
    \includegraphics[width=0.31\linewidth]{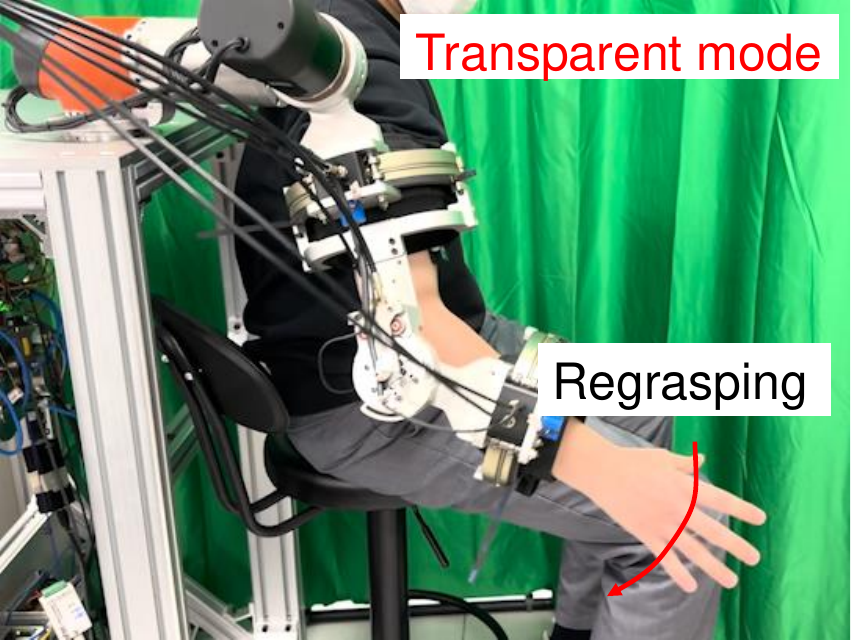}}
\subfigure[]{
    \includegraphics[width=1\linewidth]{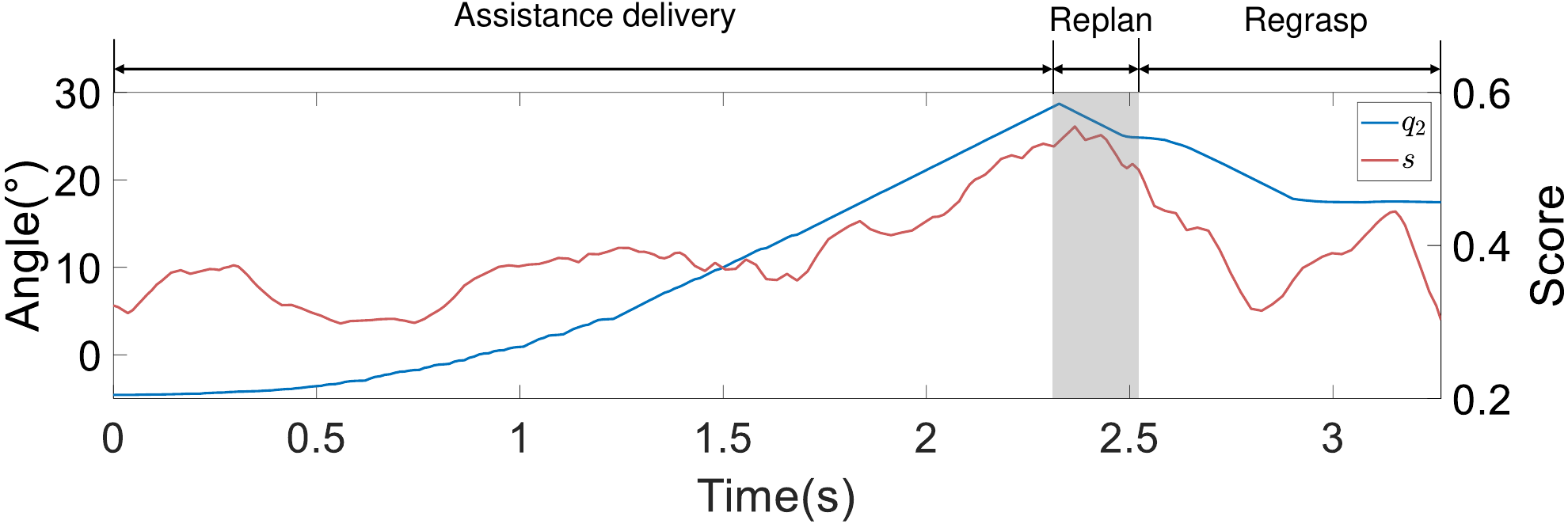}}
    \vspace{-0.5cm}
\caption{(a)–(c) Snapshots of object dropping during transportation; (d) Joint 2 angle and anomaly score in different stages. The shaded region indicates the replanning process triggered when the score exceeds the threshold, enabling the exoskeleton to operate in a transparent mode for regrasping.}\label{vary_weight}
\vspace{-0.6cm}
\end{figure}

\begin{table}[t!]
\centering
\caption{Average Assistance in different Configurations}
\begin{tblr}{
  cell{1}{1} = {r=2}{},
  cell{1}{2} = {r=2}{},
  cell{1}{3} = {c=4}{},
  vlines,
  hline{1,6} = {-}{0.08em},
  hline{2} = {3-6}{},
  hline{3-5} = {-}{},
}
Object   & (Spd,Imp) & Assistance ($N\cdot m$) &         &         &         \\
         &           & Joint 1    & Joint 2 & Joint 3 & Joint 4 \\
Water    & (0,0)     & -1.31      & -2.02   & 2.13    & -0.28   \\
Apple    & (1,1)     & 1.98       & 2.68    & -       & 1.17    \\
Dumbbell & (0,1)     & -          & 4.47    & -       & 3.01    
\end{tblr}
\label{semantic_exp}
\vspace{-0.5cm}
\end{table}

\subsection{Assistance Delivery}

When the impedance mode is activated with $\bm C_d=30\bm I_3$ and $\bm K_d=50\bm I_3$, the exoskeleton assists according to the configured parameters.
Specifically, $w$ sets impedance and parameters $t_f$ with $\bar{\dot{\bm q}}_d$ control speed by adjusting motion duration and employing specific joint velocity constraints through online refinement. 
This process was validated in an arm-raising and lowering motion involving Joint 2, as shown in Fig.~\ref{refine_check}. 
This approach ensured compliance with the $30^\circ/s$ limit, completing the motion with reduced tracking error ($1.73^\circ$ vs. $1.87^\circ$) compared to the unrefined trajectory.
% Without refinement, the joint velocity exceeds the predefined speed limit of $30^\circ/s$. Conversely, online refinement dynamically adjusts the desired trajectory based on velocity constraints, ensuring that the movement is completed while adhering to the speed limit and achieving smaller tracking errors (average errors: $1.73^\circ$ vs. $1.87^\circ$).

To evaluate the proposed framework’s effectiveness in delivering semantic-aware assistance across tasks, we conducted a grasping experiment involving three semantically distinct objects, a water bottle, an apple, and a dumbbell, with three voluntary participants, no personal data collection, and full adherence to standard ethical guidelines. Target poses were predefined, and joint engagement varied according to the task context. As summarized in Table~\ref{semantic_exp}, the system successfully derived appropriate assistance configurations from semantic information during task planning: objects such as the apple and dumbbell, suitable for higher impedance, received substantial joint assistance, whereas the water bottle, requiring lower impedance and speed, elicited reduced or even negative assistance in certain joints, reflecting the wearer’s intentional trajectory slowdown to ensure stable manipulation while assistance was maintained in others.

We further evaluated the task replanning mechanism by monitoring the anomaly score in scenarios involving object drops during transportation. The anomaly detector, trained on interactive data collected while the exoskeleton was freely moved by a wearer, triggered replanning when the score exceeded a 0.5 threshold. As illustrated in Fig.~\ref{vary_weight}, the exoskeleton initially operated in impedance control mode to assist in grasping a water bottle, during which the anomaly score remained low, indicating normal operation. Upon object drop, the score significantly increases, initiating the replanning process.
% We further evaluated the task replanning mechanism using the anomaly score in response to object dropping during transportation.
% The anomaly detector was trained on interactive information derived from our own dataset, where the exoskeleton was freely moved by a wearer. The replanning was triggered when the anomaly score exceeded a threshold of 0.5.
% As shown in Fig.~\ref{vary_weight}, the exoskeleton initially operates in impedance control mode, assisting the wearer in grasping a water bottle. 
% During this process, the anomaly score remains low, indicating normal assistance. 
% However, when the object is dropped, the score significantly increases, triggering a replanning event. 
{The exoskeleton switches to transparent mode, allowing the wearer's intention to intervene and ultimately facilitating a successful regrasping, while simultaneously prompting the LLM in parallel to generate a new task plan.
During this intervention phase, the impedance controller parameters are not applied.}
%The exoskeleton then communicates with the LLM to handle the anomaly through replanning while simultaneously switching to transparent mode, allowing the wearer's intention to intervene and ultimately facilitating a successful regrasping.

\section{Conclusions}
This paper presents a semantic-aware framework for exoskeletons that integrates user intent and safety through anomaly detection and online trajectory refinement. Leveraging LLMs, it enables automated task planning from simple descriptions, enhancing adaptability in homecare settings. Experiments demonstrate effective cognitive alignment, safe execution, and flexible integration of new movement primitives.
{Future work will explore voice control, larger-scale experiments, a broader set of motion primitives, refinement on the granularity of tuning, and integration of external sensors (e.g., cameras, IMUs) to improve intent understanding and task adaptability.}

{\small
\bibliographystyle{ref/IEEEtran}
\bibliography{ref/ref}
}

\end{document}